\documentclass[10pt,twocolumn,letterpaper]{article}

\usepackage[pagenumbers]{cvpr} 

\usepackage{graphicx}
\usepackage{amsmath}
\usepackage{amssymb}
\usepackage{booktabs}
\usepackage[accsupp]{axessibility}

\usepackage[pagebackref,breaklinks,colorlinks]{hyperref}

\usepackage[capitalize]{cleveref}
\crefname{section}{Sec.}{Secs.}
\Crefname{section}{Section}{Sections}
\Crefname{table}{Table}{Tables}
\crefname{table}{Tab.}{Tabs.}

\def\cvprPaperID{10289} 
\def\confName{CVPR}
\def\confYear{2023}

\usepackage[table]{xcolor}
\usepackage{algpseudocode}
\algrenewcommand\algorithmicrequire{\textbf{Input:}}
\algrenewcommand\algorithmicensure{\textbf{Output:}}
\usepackage{amsmath,amssymb} 
\usepackage{tabu}
\usepackage{amstext} 
\usepackage{booktabs}
\usepackage[tight-spacing=true]{siunitx}
\usepackage[normalem]{ulem}
\usepackage{etoolbox}
\usepackage{array}
\usepackage{makecell}
\usepackage{enumitem}

\usepackage{float}
\usepackage{multirow}
\usepackage{bm}

\def\eg{\emph{e.g}\onedot} 
\def\ie{\emph{i.e}\onedot}

\usepackage{booktabs} 
\usepackage{subcaption}
\usepackage[accsupp]{axessibility}  
\usepackage{amsthm}
\usepackage{thmtools, thm-restate}

\usepackage{pifont}
\usepackage{booktabs}
\usepackage{siunitx}

\renewrobustcmd{\bfseries}{\fontseries{b}\selectfont}

\usepackage{amsmath,mathtools,calc}
\usepackage{xkeyval}
\makeatletter
\newlength{\LP@lh@boxwidth}      
\newlength{\LP@lh@boxsep}        
\newcommand{\LP@lh@content}{}
\newcommand{\LP@lh@tagname}{}    
\define@key[LP]{lhbox}{boxsep}[2em]{\setlength{\LP@lh@boxsep}{#1}}
\define@key[LP]{lhbox}{content}[]{\renewcommand{\LP@lh@content}{#1}}
\define@key[LP]{lhbox}{tagname}[]{\renewcommand{\LP@lh@tagname}{#1}}
\newenvironment{tagblock}[1][]%
                         {\setkeys[LP]{lhbox}{tagname,boxsep,#1}\relax%
                          \settowidth{\LP@lh@boxwidth}{\tagform@\LP@lh@tagname}%
                          \noindent%
                          \parbox{\LP@lh@boxwidth}{\begin{align}\tag{\LP@lh@tagname}\LP@lh@content\end{align}}%
                          \hspace*{\LP@lh@boxsep}%
                          \begin{minipage}{\linewidth-\LP@lh@boxwidth-\LP@lh@boxsep}}%
                         {~\end{minipage}}

\begin{document}
\title{Bias Mimicking: A Simple Sampling Approach for Bias Mitigation}

\author{$\text{Maan Qraitem}^{1}$, $\text{Kate Saenko}^{1,2}$, $\text{Bryan A. Plummer}^1$\\
$^1\text{Boston University}$ $\text{   }$ $^2\text{MIT-IBM Watson AI Lab}$\\
\{\tt\small mqraitem,saenko,bplum\}@bu.edu\\
}

\maketitle

\begin{abstract}
Prior work has shown that Visual Recognition datasets frequently underrepresent bias groups $B$ (\eg Female) within class labels $Y$ (\eg Programmers). This dataset bias can lead to models that learn spurious correlations between class labels and bias groups such as age, gender, or race. Most recent methods that address this problem require significant architectural changes or additional loss functions requiring more hyper-parameter tuning. Alternatively, data sampling baselines from the class imbalance literature (\eg Undersampling, Upweighting), which can often be implemented in a single line of code and often have no hyperparameters, offer a cheaper and more efficient solution. However, these methods suffer from significant shortcomings. For example, Undersampling drops a significant part of the input distribution per epoch while Oversampling repeats samples, causing overfitting. To address these shortcomings, we introduce a new class-conditioned sampling method: Bias Mimicking. The method is based on the observation that if a class $c$ bias distribution, \ie $P_D(B|Y=c)$ is mimicked across every $c^{\prime}\neq c$, then $Y$ and $B$ are statistically independent. Using this notion, BM, through a novel training procedure, ensures that the model is exposed to the entire distribution per epoch without repeating samples. Consequently, Bias Mimicking improves underrepresented groups' accuracy of sampling methods by 3\% over four benchmarks while maintaining and sometimes improving performance over nonsampling methods. Code: \url{https://github.com/mqraitem/Bias-Mimicking}
\end{abstract}
\begin{figure}[th!]
\centering
\includegraphics[width=0.95\linewidth]{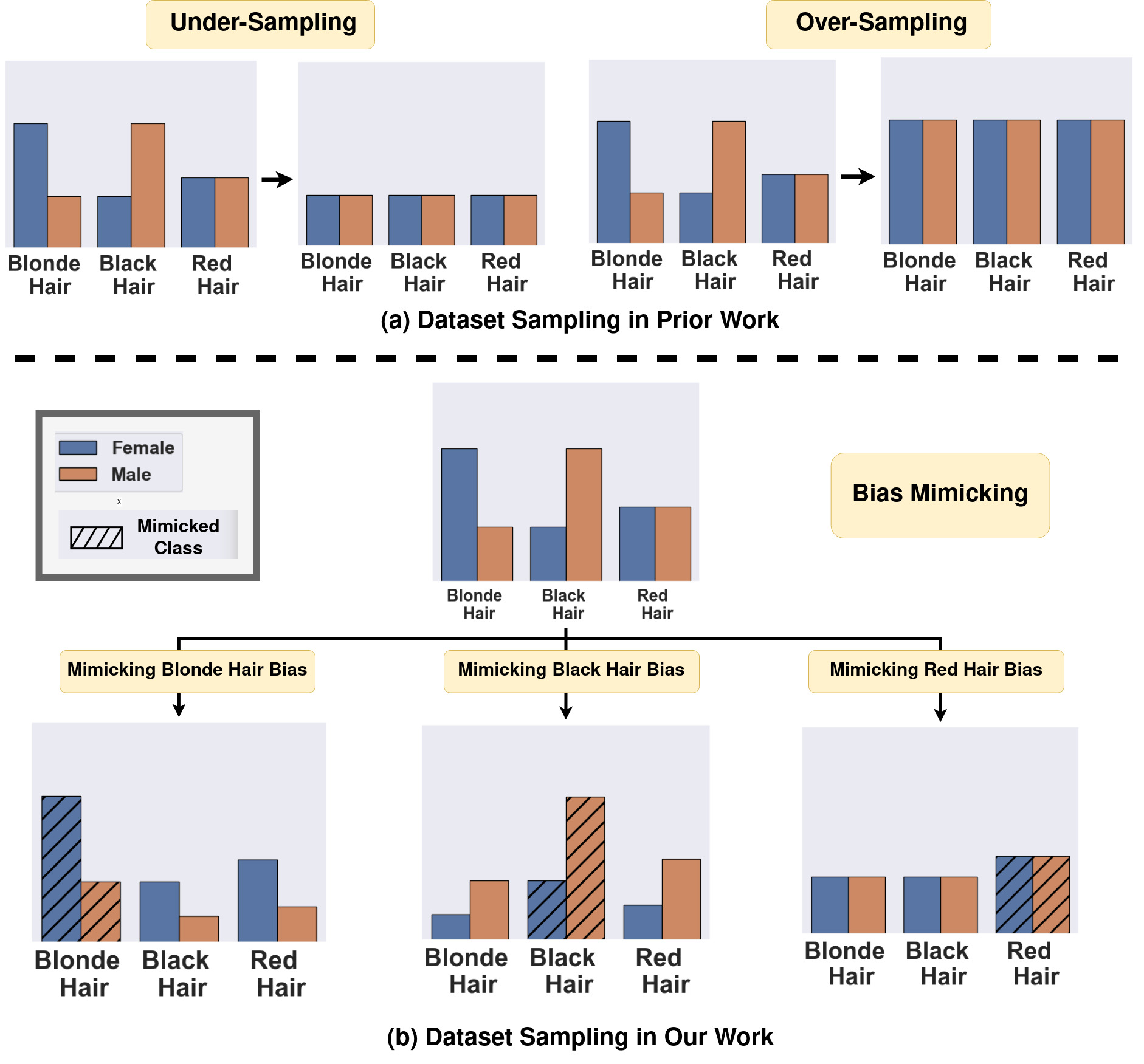}
\caption{Comparison of sampling approaches for mitigating bias of class labels $Y$ (Hair Color) toward sensitive group labels $B$ (Gender).  \textbf{(a)} illustrates Undersampling/Oversampling methods that drop/repeat samples respectively from a dataset $D$ per epoch and thus ensure that $P_D(Y | B) = P_D(Y)$. However, dropping samples hurt the model's predictive performance, and repeating samples can cause overfitting with over-parameterized models like neural nets \cite{Wang_2020_CVPR}. \textbf{(b)} shows our Bias Mimicking approach which subsamples $D$ and produces three variants. Each variant, denoted as $d_c \subset D$, preserves class $c$ samples (\ie mimicked class) and mimics the bias of class $c$ in each $c^{\prime}\neq c$. This mimicking process, as we show in our work, ensures that $P_{d_c}(Y | B) = P_{d_c}(Y)$. Moreover, by using each $d_c$ separately to train the model, we expose it to all the samples in $D$ per epoch, and since we do not repeat samples in each $d_c$, our method is less prone to overfitting.}
\label{fig:figure_1}
\vspace{-3mm}
\end{figure}

\section{Introduction}
\label{sec:intro}

Spurious predictive correlations have been frequently documented within the Deep Learning literature \cite{revise_olga,zhao2021captionbias}. These correlations can arise when most samples in class a $c$  (\eg blonde hair) belong to a bias group $s$ (\eg female). Thus, the model might learn to predict classes by using their membership to their bias groups (\eg more likely to predict blonde hair if a sample is female). Mitigating such spurious correlations (Bias) involves decorrelating the model's predictions of input samples from their membership to bias groups. Previous research efforts have primarily focused on model-based solutions. These efforts can be mainly categorized into two directions 1) ensemble-based methods \cite{Wang_2020_CVPR}, which introduce separate prediction heads for samples from different bias groups 2) methods that introduce additional bias regularizing loss functions and require additional hyper-parameter tuning  \cite{hong2021unbiased,8953715,ryu2018inclusivefacenet,Tartaglione_2021_CVPR,Sagawa2020DistributionallyRN}.

Dataset resampling methods, popular within the class imbalance literature \cite{Japkowicz02theclass,BUDA2018249,pmlr-v119-sagawa20a,pmlr-v81-buolamwini18a}, present a simpler and cheaper alternative. They do not require hyperparameter tuning or extra model parameters. Therefore, they are faster to train. Moreover, as illustrated in Figure \ref{fig:figure_1}(a), they can be extended to Bias Mitigation by considering the imbalance within the dataset subgroups rather than classes. Most common of these methods are Undersampling \cite{Japkowicz02theclass,BUDA2018249,pmlr-v119-sagawa20a} and Oversampling \cite{Wang_2020_CVPR}. They mitigate class imbalance by altering the dataset distribution through dropping/repeating samples, respectively. Another similar solution is Upweighting \cite{upweight2,pmlr-v97-byrd19a}, which levels each sample contribution to the loss function by appropriately weighting its loss value. However, these methods suffer from significant shortcomings. For example, Undersampling drops a significant portion of the dataset per epoch, which could harm the models' predictive capacity. Moreover, Upweighting can be unstable when used with stochastic gradient descent \cite{DBLP:conf/iclr/AnYZ21}. Finally, models trained with Oversampling, as shown by \cite{Wang_2020_CVPR}, are likely to overfit due to being exposed to repetitive sample copies. 

To address these problems, we propose Bias Mimicking (BM): a class-conditioned sampling method that mitigates the shortcomings of prior work. As shown in Figure \ref{fig:figure_1}(b), given a dataset $D$ with a set of three classes $C$, BM subsamples $D$ and produces three different variants. Each variant, $d_c \subset D$ retains every sample from class $c$ while subsampling each $c^{\prime} \neq c$  such that $c^{\prime}$ bias distribution, \ie $P_{d_c }(B | Y = c^{\prime})$,  mimics that of $c$. For example, observe $d_{\text{Blonde Hair}}$ in Figure \ref{fig:figure_1}(b) bottom left. Note how the bias distribution of class "Blonde Hair" remains the same while the bias distributions of "Black Hair" and "Red Hair" are subsampled such that they mimic the bias distribution of "Blonde Hair". This mimicking process decorrelates $Y$ from $B$ since $Y$ and $B$ are now statistically independent as we prove in Section \ref{sec:bias_mimic}. 

The strength of our method lies in the fact that $d_c$ retains class $c$ samples while at the same time ensuring $P_{d_c}(Y|B) = P_{d_c}(Y)$ in each $d_c$. Using this result, we introduce a novel training procedure that uses each distribution separately to train the model. Consequently, the model is exposed to the entirety of $D$ since each $d_c$ retains class $c$ samples. Refer to Section \ref{sec:bias_mimic} for further details. Note how our method is fundamentally different from Undersampling. While Undersampling also ensures  statistical independence on the dataset level, it subsamples every subgroup. Therefore, the training distribution per epoch is a smaller portion of the total dataset $D$. Moreover, our method is also different from Oversampling since each $d_c$ does not repeat samples. Thus we reduce the risk of overfitting. 

In addition to proposing Bias Mimicking, another contribution of our work is providing an extensive analysis of sampling methods for bias mitigation. We found many sampling-based methods were notably missing in the comparisons used in prior work \cite{Tartaglione_2021_CVPR, Wang_2020_CVPR, hong2021unbiased}.
Despite their shortcomings, we show that Undersampling and Upweighting are surprisingly competitive on many bias mitigation benchmarks. Therefore, this emphasizes these methods' importance as an inexpensive first choice for mitigating bias. However, in cases where these methods are ineffective, Bias Mimicking bridges the performance gap and achieves comparable performance to nonsampling methods. Finally, we thoroughly analyze our approach's behavior through two experiments. First, we verify the importance of each $d_c$ to the model's predictive performance in Section \ref{sec:analysis_dy}. Second, we investigate our method's sensitivity to the mimicking condition in Section \ref{sec:exp_analysis}. Both experiments showcase the importance of our design in mitigating bias. 

Our contributions can be summarized as: 
\begin{itemize}[nosep,leftmargin=*]
    \item We show that simple sampling methods can be competitive on some benchmarks when compared to non sampling state-of-the-art approaches. 
    \item We introduce a novel resampling method: Bias Mimicking that bridges the performance gap between sampling and nonsampling methods; it improves the average under-represented subgroups accuracy by $>3\%$ compared to other sampling methods.  
    \item We conduct an extensive empirical analysis of Bias Mimicking that details the method's sensitivity to the Mimicking condition and uncovers insights about its behavior. 
\end{itemize}

\begin{figure*}[th!]
    \centering
    \includegraphics[width=0.9\linewidth]{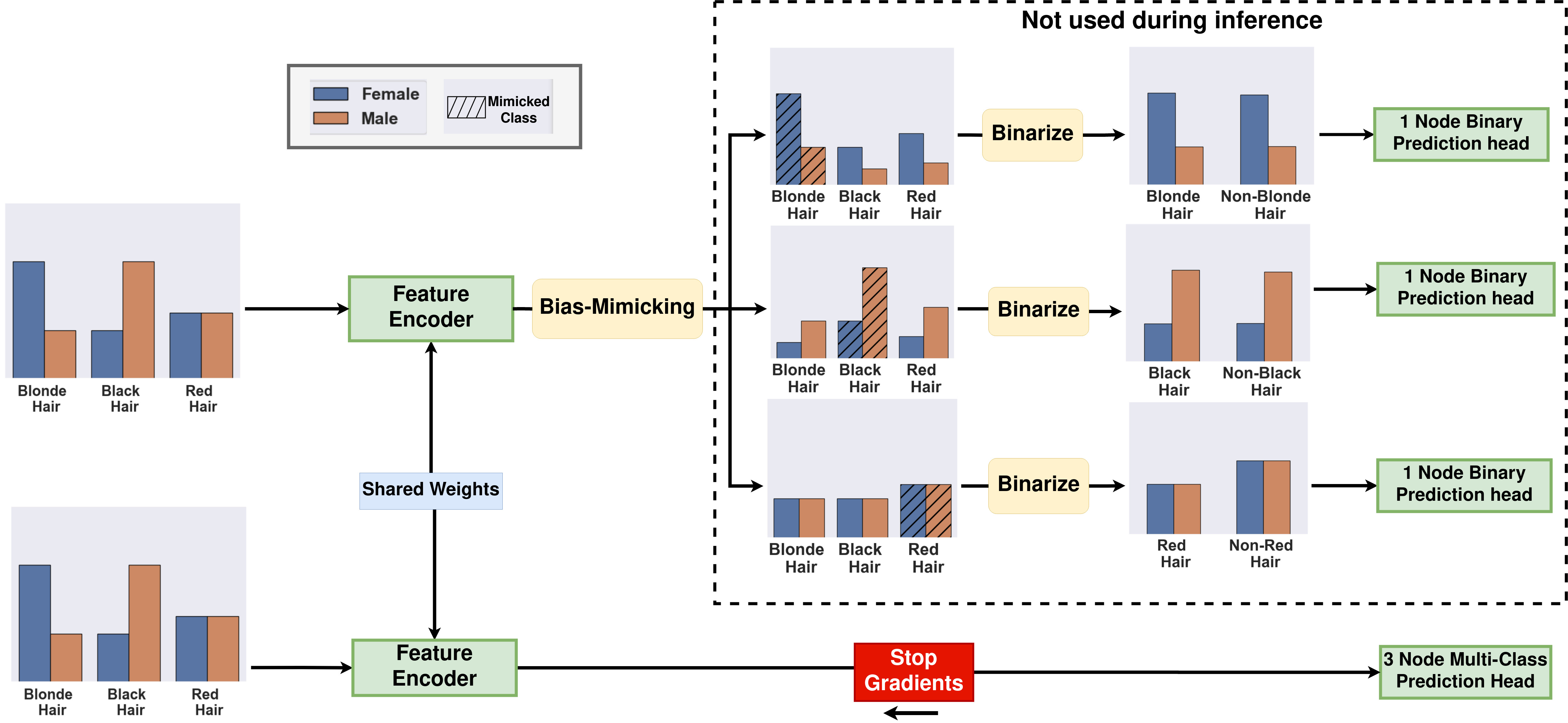}
    \caption{Given a dataset $D$ with three classes, Bias Mimicking subsamples $D$ and produces three different distributions. Each distribution $d_c \subset D$ retains class $c$ samples while subsampling each $c^{\prime} \neq c$ to ensure the bias distribution of $c$, \ie $P_{d_{c}}(B|Y=c)$ is mimicked in each $c^{\prime}$. To train an image classification model on the resulting distributions, we binarize each and feed it through a dedicated binary prediction head. Alternatively, we could dedicate a multi-class prediction head for each distribution. However, this alternative will introduce significantly more parameters. Put together, the binary prediction heads are equivalent  \#parameters wise to a model that uses one multi-class prediction head. To perform inference, we can not use the scores from the binary predictors because they may not be calibrated with respect to each other. Therefore, we train a multi-class prediction head using the debiased feature encoder from Bias Mimicking and freeze the gradient from flowing back to ensure the bias is not relearned. } 
    \label{fig:figure_2}
    \vspace{-2mm}
\end{figure*}
\section{Related Work}

\smallskip
\noindent\textbf{Documenting Spurious Correlations.} Bias in Machine Learning can manifest in many ways. Examples include class imbalance \cite{Japkowicz02theclass}, historical human biases \cite{Suresh2019AFF}, evaluation bias \cite{pmlr-v81-buolamwini18a}, and more. For a full review, we refer the reader to \cite{10.1145/3457607}. In our work, we are interested in model bias that arises from spurious correlations. A spurious correlation results from underrepresenting a certain group of samples (\eg samples with the color red) within a certain class (\eg planes). This leads the model to learn the false relationship between the class and the over-represented group. Prior work has documented several occurrences of this bias. For example, \cite{singh-cvpr2020,hendrycks2021nae,xiao2020noise,li2020shapetexture} showed that state-of-the-art object recognition models are biased toward backgrounds or textures associated with the object class. \cite{dontlook,clark-etal-2019-dont} showed similar spurious correlations in VQA. \cite{zhao2021captionbias}  noted concerning correlations between captions and attributes like skin color in Image-Captioning datasets. Beyond uncovering these correlations within datasets, prior work like \cite{Wang_2020_CVPR,hong2021unbiased} introduced synthetic bias datasets where they systematically assessed bias effect on model performance. Most recently, \cite{Meister2022GenderAI} demonstrates how biases toward gender are ubiquitous in the COCO and OpenImages datasets. As the authors demonstrate, these artifacts vary from low-level information (e.g., the mean value of the color channels) to the higher level (e.g., pose and location of people). 

\smallskip
\noindent\textbf{Nonsampling solutions.} In response to the documentation of dataset spurious correlations, several model-focused methods have been proposed  \cite{ryu2018inclusivefacenet,8953715,Wang_2020_CVPR,hong2021unbiased,Tartaglione_2021_CVPR,sagawa2019distributionally}. For example, \cite{Tartaglione_2021_CVPR} presents a metric learning-based method where model feature space is regularized against learning harmful correlations. \cite{Wang_2020_CVPR} surveys several existing methods, such as adversarial training, that randomize the relationship between target classes and bias groups in the feature space. They also present a new approach, domain independence, where different prediction heads are allocated for each subgroup. \cite{sagawa2019distributionally} presents GroupDRO (Distributionally Robust Neural Networks for Group Shifts), a regularization procedure that adapts the model optimization according to the worst-performing group. Most recently, \cite{hong2021unbiased} extended contrastive learning frameworks on self-supervised learning \cite{chen2020big,9157636,NEURIPS2020_d89a66c7} to mitigate bias. Our work complements these efforts by introducing a hyperparameter-free sampling algorithm that bridges the performance gap between nonsampling and sampling methods.

\smallskip
\noindent\textbf{Dataset based solutions.} In addition to model-based approaches, we can mitigate spurious correlations by fixing the training dataset distribution. Examples include Oversampling minority classes, Undersampling majority ones, and weighting the loss value of different samples to equalize the contribution of dataset subgroups. These approaches are popular within the class imbalance literature \cite{Japkowicz02theclass,BUDA2018249,pmlr-v119-sagawa20a,pmlr-v81-buolamwini18a}. However, as noted in the introduction, some of these methods have been missing in recent visual Bias Mitigation Benchmarks \cite{Wang_2020_CVPR, Tartaglione_2021_CVPR, hong2021unbiased}. Thus, we review these methods and describe their shortcomings in Section \ref{sec:prior_sampling}. Alternatively, other efforts attempt to fix the dataset distribution by introducing new samples. Examples include \cite{9423296}, where they introduce a new dataset for face recognition balanced among several race groups, and \cite{DBLP:journals/corr/abs-2012-01469}, where they used GANs to generate training data that balance the sizes of dataset subgroups. While our work is a dataset-based approach, it differs from these efforts as it does not generate or introduce new samples. Finally, also related to our work are sampling methods like REPAIR \cite{Li2019REPAIRRR}, where a function is learned to prioritize specific samples and, thus, learn more robust representations. 


\section{Sampling For Bias Mitigation}
\label{sec:methods}

In visual bias mitigation, the goal is to train a model that does not rely on spurious signals in the images when making predictions (\eg not using gender signal when predicting hair color). Formally, assume we have a dataset of image/target-classes/bias-group triplets  $(X, Y, B)$ where the images $X$ contain components denoted as $X_b$ that determine their bias group memberships $B$. Furthermore, let $C$ represents the set of possible target-classes, and $S$ represents the set of possible bias groups. We characterise a model behavior as biased when the model uses the biased image signal, \ie $X_b$, to predict $Y$. This behavior might arise when a target class $c \in C$ (\eg blonde hair) is over-represented by images that belong to one bias group $s \in S$ (\eg female) rather than distributed equally among the dataset bias groups. In other words, there exists $s \in B$ for certain class $c \in C$ such that $P_{D}(B=s | Y=c) >> \frac{1}{|S|}$ where $|S|$ denotes the number of bias groups.  For example, if most blonde hair samples were female, the model might use the image components corresponding to gender in making predictions rather than solely relying on hair color. 

Our work addresses methods that fix spurious correlations by ensuring statistical independence between $Y$ and $B$ on the dataset level, \ie $P_D(Y|B) = P_D(Y)$. In that spirit, we introduce a new sampling method: Bias Mimicking in Section \ref{sec:bias_mimic}. However, as we note in our introduction, some popular sampling solutions in class imbalance literature that could also be applied to Bias Mitigation (\eg Undersampling, Upweighting) are missing from benchmarks used in prior work. Therefore, we briefly review the missing methods in Section \ref{sec:prior_sampling} and describe how Bias Mimicking addresses their shortcomings.

\subsection{Bias Mimicking}
\label{sec:bias_mimic}
The goal of our method is to prevent the model from using the biased signal in the images, denoted as $X_b$, in making predictions $\hat{Y}$. Our approach is inspired by Sampling methods, which mitigate this problem by enforcing statistical Independence between the target labels $Y$ and the bias groups $B$ in the dataset, \ie, $P_D(Y|B)= P_D(Y)$. However, simple sampling methods like Oversampling, Undersampling, and Upweighting need to be improved as we discuss in the introduction. We address these shortcomings in our proposed sampling algorithm: Bias Mimicking. 

\smallskip
\noindent\textbf {Mimicking Distributions} The key to our algorithm is the observation that if the distribution of bias groups with respect to a class $c$; \ie, $P_D(B | Y = c)$, was the same across every $c \in C$, then $B$ is statistically independent from $Y$. For example, consider each resulting distribution from our Bias Mimicking process in Figure \ref{fig:figure_2}. Note how in each resulting distribution, the distribution of bias group "Gender" is the same across every class "Hair Color" which ensures statistical independence between "Gender" and "Hair Color" on the dataset level. Formally:

\begin{restatable}{proposition}{Firstprop}
\label{prop:1}
Given dataset $D$, target classes set $C$, bias groups set $S$, target labels $Y$,  bias groups labels $B$, and bias group $s \in S$ if $P_D(B = s | Y = i) = P_D(B = s | Y = j) \quad\forall i,j \in C$, then  $P_D(Y = c | B = s) = P_D(Y = c) \quad \forall c \in C.$
\end{restatable}
 
Note that ensuring this proposition holds for every $s \in S$ (\ie the distribution of $P_D(B|Y = c)$ is the same for each $c \in C$) implies that $P_D(Y|B) = P_D(Y)$. Proving the proposition above is a simple application of the law of total probability. Given $s \in S$, then
\begin{equation}
    P_D(B = s) = \sum_{c \in C} P_D(B = s | Y = c) P_D(Y = c)
\end{equation}
Given our assumption of bias mimicking, \ie $P_D(B = s | Y = i) = P_D(B = s | Y = j) \quad\forall i,j \in C$, then we can rewrite (1) $\forall c \in C$ as: 
\begin{align}
    P_D(B = s) & = P_D(B = s | Y = c)  \sum_{c^{\prime} \in C} P_D(Y = c^{\prime})\\ 
             &= P_D(B = s | Y = c) 
\end{align}
From here, using Bayesian probability and the result from (3),  we can write, $\forall c \in C$: 
\begin{align}
    P_D(Y=c | B=s) &= \frac{P_D(B=s | Y=c)P_D(Y=c)}{P_D(B=s)} \\ 
                 &= \frac{P_D(B=s)P_D(Y=c)}{P_D(B=s)} \\
                 &= P_D(Y=c)
\end{align}
We use this result to motivate a novel way of subsampling the input distribution that ensures the model is exposed to every sample in the dataset. To that end, for every class $c \in C$, we produce a subsampled version of the dataset $D$ denoted as $d_c \subset D$. Each $d_c$ preserves its respective class $c$ samples, \ie, all the samples from class $c$ remain in $d_c$,  while subsampling every $c^{\prime} \neq c$ such that the bias distribution in class $c$; \ie $P_{d_c}(B | Y=c)$, is mimicked in each $c^{\prime}$. Formally: 
\begin{equation}
    \label{eq:biasmimick}
    P_{d_c}(B=s | Y = c) = P_{d_c}(B = s | Y = c^{\prime}) \quad \forall s \in S.
\end{equation}

Indeed, as Figure \ref{fig:figure_2} illustrates, a dataset of three classes will be subsampled in three different ways. In each version, a class bias is mimicked across the other classes. Note that there is not unique solution for mimicking distributions. However, we can constrain the mimicking process by ensuring that we retain the most number of samples in each subsampled $c^{\prime} \neq c$. We enforce this constraint naturally through a linear program. Denote the number of samples that belong to class $c^{\prime}$ as well as the bias group $s$ as $l^{c^{\prime}}_s$. Then we seek to optimize each $l^{c^{\prime}}_s$ as follows: 

\begin{tagblock}[tagname={P.1},content={\label{linear_program}}]
    \begin{flalign}\notag
      \max \quad &\sum_{s} l^{c^{\prime}}_s \\
      \text{s.t.\quad} &  l^{c^{\prime}}_s  \leq  |D_{c^{\prime}, s}| &   & s \in S\\
                      &  \frac{l^{c^{\prime}}_s}{\sum_{s} l^{c^{\prime}}_s }  =  P_{D}(B=s | Y = c)  &   & s \in S
    \end{flalign}
\end{tagblock}

\noindent where $|D_{c,s}|$ represents the number of samples that belong to both class $c$ and bias group $s$. This linear program returns the distribution with most number of retained samples while ensuring that the mimicking condition holds. 

\noindent\textbf {Training with Bias Mimicking} We use every resulting $d_c \subset D$ to train the model. However, we train our model such that it sees each $d_c$ through a different prediction head to ensure that the loss function does not take in gradients of repetitive copies, thus risking overfitting. A naive way of achieving this is to dedicate a multi-class prediction head for each $d_c$. However, this choice introduces ($|C|-1$) additional prediction heads compared to a model that uses only one. To avoid this, we binarise each $d_c$ and then use it to train a one-vs-all binary classifier $\text{BP}_{c}$ for each $c$. Each head is trained on image-target pairs from  its corresponding distribution $d_c$ as Figure \ref{fig:figure_2} demonstrates. Note that each binary predictor is simply one output node. Therefore, the binary predictors introduce no extra parameters compared to a multi-class head. Finally, given our setup, a prediction head might see only a small number of samples at certain iterations. We mitigate this problem by accumulating the gradients of each prediction head across iterations and only backpropagating them once the classifier batch is full. 

\noindent\textbf {Inference with Bias Mimicking} Using the binary classifiers during inference is challenging since each was trained on different distributions, and their scores, therefore, may not be calibrated. However, Bias Mimicking minimizes the correlation between $Y$ and $B$ within the feature space. Thus, we exploit this fact by training a multi-class prediction head over the feature space using the original dataset distribution. However, this distribution is biased; therefore, we stop gradients from flowing into the model backbone when training that layer to ensure that the bias is not learned again in the feature space, as Figure \ref{fig:figure_2} demonstrates.

\noindent\textbf {Cost Analysis}  Unlike prior work \cite{hong2021unbiased,Tartaglione_2021_CVPR}, bias mimicking involves no extra hyper-parameters. The debiasing is automatic by definition. This means that the hyper-parameter search space of our model is smaller than other bias mitigation methods like \cite{hong2021unbiased,sagawa2019distributionally}. Consequently, our method is cheaper and more likely to generalize to more datasets as we show in Section \ref{sec:main_results}. Moreover, note that the additional input distributions $d_c$ do not result in longer epochs; we make one pass only over each sample $x$ during an epoch and apply its contribution to the relevant $BP_y$(s). Our only additional cost is solving the linear program \ref{linear_program} and training the multi-class prediction head. However, training the linear layer is simple, fast, and cheap since it is only \textit{one} linear layer. Moreover, the linear program is solved through modern and efficient implementations that take seconds and only needs to be done once for each dataset.

\subsection{Simple Sampling Methods}
\label{sec:prior_sampling}
As discussed in the introduction, the class imbalance literature is rich with dataset sampling solutions \cite{Japkowicz02theclass,BUDA2018249,pmlr-v119-sagawa20a,pmlr-v81-buolamwini18a,smote}. These solutions address class imbalance by balancing the distribution of classes. Thus, they can be extended to Bias Mitigation by balancing groupings of classes and bias groups (\eg group male-black hair). Prior work in Visual Bias Mitigation has explored one of these solutions: Oversampling \cite{Wang_2020_CVPR}. However, other methods like Undersampling \cite{Japkowicz02theclass,BUDA2018249,pmlr-v119-sagawa20a}  and Upweighting \cite{upweight2,pmlr-v97-byrd19a} are popular alternatives to Oversampling. Both methods, however, have not been benchmarked in recent Visual Bias Mitigation work \cite{Tartaglione_2021_CVPR,hong2021unbiased}. We review these sampling solutions below and note how Bias Mimicking addresses their shortcomings. 

\begin{table*}[t!]
\sisetup{table-number-alignment=center}
\sisetup{
  table-align-uncertainty=true,
  separate-uncertainty=true,
  detect-all,
  detect-weight=true,
  detect-shape=true, 
  detect-mode=true,
}
\setlength{\tabcolsep}{0.2em}
\renewcommand{\arraystretch}{1.1}
\centering
\begin{tabular}{ll|S[table-format=2.1(1),detect-all]|S[table-format=2.1(1),detect-all]S[table-format=2.1(1),detect-all]S[table-format=2.1(1),detect-all]S[table-format=2.1(1),detect-all]|S[table-format=2.1(1),detect-all]S[table-format=2.1(1),detect-all]S[table-format=2.1(1),detect-all]S[table-format=2.1(1),detect-all]S[table-format=2.1(1),detect-all]S[table-format=2.1(1),detect-all]}
\toprule
 &  &  & \multicolumn{4}{c|}{Non Sampling Methods} & \multicolumn{4}{c}{Sampling Methods} \\ \midrule
 &  & {Vanilla} & {Adv \cite{Hoffman2015SimultaneousDT}} & {G-DRO \cite{Sagawa2020DistributionallyRN}} & {DI\cite{Wang_2020_CVPR}} & {BC+BB}\cite{hong2021unbiased} & {OS \cite{Wang_2020_CVPR}} & {UW \cite{pmlr-v97-byrd19a}} & {US\cite{Japkowicz02theclass}} & {BM (ours)} \\ \midrule
UTK-Face & UA & 72.8\pm0.2 & 70.2\pm0.1 & 74.2\pm0.9 & 75.5\pm1.1 & \underline{\tablenum{78.9\pm0.5}} & 76.6\pm0.3 & 78.8\pm1.2 & 78.2\pm1.0 & \underline{\tablenum{79.7\pm 0.4}} \\
Age & BC & 47.1\pm0.3 & 44.1\pm1.2 & \underline{\tablenum{75.9\pm2.9}} & 58.8\pm3.0\ & 71.4\pm2.9 & 58.1\pm1.2\ & 77.2\pm3.8 & 69.8\pm6.8 & \underline{\tablenum{79.1\pm 2.3}} \\ \midrule
UTK-Face & UA & 88.4\pm0.2 & 86.1\pm0.5 & 90.8\pm0.3 & 90.7\pm0.1 & \underline{\tablenum{91.4\pm0.2}} & \underline{\tablenum{91.3\pm0.6}} & 89.7\pm0.6 & 90.8\pm0.2 & 90.8\pm 0.2\\
Race & BC & 80.8\pm0.3 & 77.1\pm1.3 & 90.2\pm0.3 & \underline{\tablenum{90.9\pm0.4}} & \underline{\tablenum{90.6\pm0.5}} & 90.0\pm0.7\ & 89.2\pm0.4 & 89.3\pm0.6 & \underline{\tablenum{90.7\pm 0.5}} \\ \midrule
CelebA & UA & 82.4\pm1.3 & 82.4\pm1.3 & 90.4\pm0.4 & \underline{\tablenum{90.9\pm0.5}} &90.4\pm0.2 & 88.1\pm0.5 & \underline{\tablenum{91.6\pm0.3}} & 91.1\pm0.2 & 90.8\pm0.4 \\
Blonde & BC & 66.3\pm2.8 & 66.3\pm2.8 & \underline{\tablenum{89.4\pm0.5}} & 86.1\pm0.8 & 86.5\pm0.5 & 80.1\pm1.2 & \underline{\tablenum{88.3\pm0.5}} & \underline{\tablenum{88.5\pm1.8}} & 87.1\pm0.6 \\ \midrule
\multirow{2}{*}{CIFAR-S} & UA & 88.7\pm0.1 & 81.8\pm2.5 & 89.1\pm0.2 & \underline{\tablenum{92.1\pm0.0}} & 90.9\pm0.2 & 87.8\pm0.2 & 86.5\pm0.5 & 88.2\pm0.4 & \underline{\tablenum{91.6\pm0.1}} \\
 & BC & 82.8\pm0.1 & 72.0\pm0.2 & 88.0\pm0.2 & \underline{\tablenum{91.9\pm0.2}} & 89.5\pm0.7 & 82.5\pm0.3 & 80.0\pm0.7 & 83.7\pm1.2 & \underline{\tablenum{91.1\pm 0.1}} \\ \midrule
\multirow{2}{*}{Average} & UA & 83.0 \pm 0.4 & 80.1\pm1.1 & 86.1\pm0.4 & 87.3\pm0.4 & \bfseries 87.9\pm0.3 & 85.9\pm0.4 & 86.6\pm0.6 & 87.0\pm0.4 & \bfseries 88.2\pm0.3 \\
 & BC & 69.2 \pm 0.8 & 64.8\pm1.4 & \bfseries 85.8\pm1.0 & 81.9\pm1.1 & 84.5\pm1.1 & 77.6\pm0.8 & 83.6\pm1.3 & 82.8\pm2.6 & \bfseries 87.0\pm0.9 \\ \bottomrule
 \end{tabular}
\caption{\textbf{Results} compare methods Adversarial training  (Adv) \cite{Kim_2019_CVPR}, GroupDRO (G-DRO) \cite{Sagawa2020DistributionallyRN},  Domain Independence DI \cite{Wang_2020_CVPR}, Bias Contrastive and Bias-Contrastive and Bias-Balanced Learning (BC+BB) \cite{hong2021unbiased}, Undersampling (US) \cite{Japkowicz02theclass}, Upweighting (UW) \cite{pmlr-v97-byrd19a}, and Bias Mimicking (BM, ours), on the CelebA, UTK-face, and CIFAR-S dataset. Methods are evaluated using Unbiased Accuracy \cite{hong2021unbiased} (UA), Bias-conflict \cite{hong2021unbiased} (BC). Given the methods' grouping: Sampling/Non Sampling, the \underline{Underlined numbers} indicate the best method per group on each dataset while \textbf{bolded numbers} indicate the best method per group on average. See Section \ref{sec:main_results} for discussion. }
\label{tb:main-results}
\end{table*}

\noindent\textbf {Undersampling} drops samples from the majority classes until all classes are balanced. We can extend this solution to Bias Mitigation by dropping samples to balance dataset subgroups where a subgroup includes every sample that shares a class $c$ and bias group $s$, \ie,  $D_{c, s} = \{(x, y, s) \in D \text{ s.t }  Y = c, B = s \} $. Then, we drop samples from each subgroup until each has a size equal to $\min\limits_{c, s} |D_{c, s}|$. Thus, in each epoch, the number of samples the model can see is limited by the size of the smallest subgroup. While we can mitigate this problem by resampling the distribution each epoch, the model, nevertheless, has to be exposed to repeated copies of the minority subgroup every time it sees new samples from the majority subgroups, which may compromise the predictive capacity of our model, as our experimental results demonstrate. \textit{Bias Mimcking} addresses this shortcoming by using each $d_c \subset D$, thus ensuring that the model is exposed to all of $D$ every epoch. 

\noindent\textbf{Oversampling}  solves class imbalance by repeating copies of samples to balance the number of samples in each class. Similarly to Undersampling, it can also be easily extended to Bias Mitigation by balancing samples across sensitive subgroups. In our implementation, we determine the maximum size subgroup (\ie $\max\limits_{c, s} |D_{c, s}|$ where $D_{c, s} = \{(x, y, s) \in D \text{ s.t }  Y = c, B = s \} $) and then repeat samples accordingly in every other subgroup. However, repeating samples as \cite{Wang_2020_CVPR} shows, cause overfitting with overparameterized models like neural nets. \textit{Bias Mimicking addresses} this problem by not repeating samples in each subsampled version of the dataset $d_c \subset D$. 

\noindent\textbf {Upweighting} Upweighting levels the contribution of different samples to the loss function by multiplying its loss value by the inverse of the sample's class frequency. We can extend this process to Bias Mitigation by simply considering subgroups instead of classes. More concretely, assume model weights $w$, sample $x$, class $c$, bias $s$, and subgroup  $D_{c, s} = \{(x, y, s) \in D \text{ s.t }  Y = c, B = s \} $, then the model optimizes:
\begin{equation*}
    L = E_{x, c, s} \Big[\frac{1}{p_D(x\in D_{c, s})} l(x, c ; w)\Big]
\end{equation*}

A key shortcoming of Upweighting is its instability when used with stochastic gradient descent \cite{DBLP:conf/iclr/AnYZ21}. Indeed, we demonstrate this problem in our experiments in Section \ref{sec:exp} where Upweighting does not work well on some datasets. \textit{Bias Mimicking addresses} this problem by not using weights to scale the loss function.

\section{Experiments}
\label{sec:exp}

We report our method performance on four total benchmarks: three binary and one multi-class classification benchmark in Section \ref{sec:main_results}. In addition to reporting our method results, we include vanilla sampling methods, namely Undersampling, Upweighting, and Oversampling, which, as noted in our introduction, have been missing from recent Bias Mitigation work \cite{hong2021unbiased,Wang_2020_CVPR,sagawa2019distributionally}. Furthermore, we report the averaged performance overall benchmarks to highlight methods that generalize well to all datasets. We then follow up our results with two main experiments that analyze our method's behavior. The first experiment in Section \ref{sec:analysis_dy} analyzes the contribution of each subsampled version of the dataset: $d_c$ to model performance. The second experiment in Section \ref{sec:exp_analysis} is a sensitivity analysis of our method to the mimicking condition. 

\subsection{Main Results}
\label{sec:main_results}

\begin{table*}[t!]
\small
\renewcommand{\arraystretch}{1.0}
\setlength{\tabcolsep}{1.8pt}
    \centering
    \begin{subtable}{.32\linewidth}
    \sisetup{
      table-align-uncertainty=true,
      separate-uncertainty=true,
      detect-all,
    }
    \centering 
        \begin{tabular}{c S[table-format=2.1,detect-weight=true] S[table-format=2.1,detect-weight=true] S[table-format=2.1(2),detect-weight=true]}
            \toprule
             & UA$_{c_1}$ & UA$_{c_2}$  & UA \\
            \midrule
            {$(d_{c_1})$}  & 82.5  & 74.2   &  78.4 \pm 0.5  \\
            {$(d_{c_2})$}  & 73.1 & 67.9  & 70.5   \pm 2.3 \\
            {$(d_{c_1}, d_{c_2})$} & \underline{\tablenum{ 84.4}} & \underline{\tablenum{75.3}} & \bfseries 79.8   \pm 0.9 \\
            \bottomrule
        \end{tabular}
        \caption{Utk-Face/Age}
    \end{subtable}
    \begin{subtable}{.32\linewidth}
    \sisetup{
      table-align-uncertainty=true,
      separate-uncertainty=true,
      detect-all,
    }
    \centering 
        \begin{tabular}{c S[table-format=2.1,detect-weight=true] S[table-format=2.1,detect-weight=true] S[table-format=2.1(2),detect-weight=true]}
            \toprule
             & UA$_{c_1}$ & UA$_{c_2}$  & UA \\
            \midrule
            {$(d_{c_1})$}  & \underline{\tablenum{90.7}}   & 85.1   &  87.9 \pm 0.5  \\
            {$(d_{c_2})$}  & 84.8 &\underline{\tablenum{91.5}}  & 88.1   \pm 0.6  \\
            {$(d_{c_1}, d_{c_2})$} &   90.5 & 91.1 & \bfseries 90.8   \pm 0.2 \\
            \bottomrule
        \end{tabular}
         \caption{Utk-Face/Race}
    \end{subtable}
    \begin{subtable}{.32\linewidth}
    \sisetup{
      table-align-uncertainty=true,
      separate-uncertainty=true,
      detect-all,
    }
    \centering 
        \begin{tabular}{c S[table-format=2.1,detect-weight=true] S[table-format=2.1,detect-weight=true] S[table-format=2.1(2),detect-weight=true]}
            \toprule
             & UA$_{c_1}$ & UA$_{c_2}$  & UA \\
            \midrule
            {$(d_{c_1})$}  & \underline{\tablenum{91.5}}  & 90.3   &  \bfseries 90.9 \pm 0.1    \\
            {$(d_{c_2})$}  & 82.2 & \underline{\tablenum{96.5}}  & 89.3  \pm 0.1 \\
            {$(d_{c_1}, d_{c_2})$} & 88.1 & 94.2 & \bfseries 90.8  \pm 0.4 \\
            \bottomrule
        \end{tabular}
            \caption{CelebaA/Blonde}
    \end{subtable}
    \caption{We investigate the effect of each resampled version of the dataset $d_c$ on model performance using the binary classification tasks outlined in Section \ref{sec:main_results}. Thus, Bias Mimicking results in two subsampled distributions $d_{c_1}$ and $d_{c_2}$. We then train three different models: 1) a model trained on only $d_{c_1}$, 2) a model trained on only $d_{c_2}$ and a model trained on both (default choice for Bias Mimicking). We use the Unbiased Accuracy metric (UA). Furthermore, we report UA on class 1: $\text{UA}_1$ and class 2: $\text{UA}_2$ separately by averaging the accuracy over the class's relevant subgroups. Refer to section \ref{sec:analysis_dy} for discussion.}
    \label{tab:dyeffect}
    \vspace{-0.6mm}
\end{table*}

\noindent\textbf{Datasets} We report performance on three main datasets: CelebA dataset \cite{liu2015faceattributes} UTKFace dataset \cite{zhifei2017cvpr} and CIFAR-S benchmark \cite{Wang_2020_CVPR}. Following prior work \cite{hong2021unbiased,Tartaglione_2021_CVPR}, we train a binary classification model using CelebA where BlondHair is a target attribute and Gender is a bias attribute. We use \cite{hong2021unbiased} split where they amplify the bias of Gender. Note that prior work \cite{hong2021unbiased,Tartaglione_2021_CVPR} used the HeavyMakeUp attribute in their CelebA benchmark. However, during our experiments, we found serious problems with the benchmark. Refer to the supplementary for model details. Therefore, we skip this benchmark in our work. For UTKFace, we follow \cite{hong2021unbiased} and do the binary classification with Race/Age as the sensitive attribute and Gender as the target attribute. We use \cite{hong2021unbiased} split for both tasks. With regard to CIFAR-S, The benchmark synthetically introduces bias into the CIFAR-10 dataset \cite{Krizhevsky2009LearningML} by converting a subsample of images from each target class to a gray-scale. We use \cite{Wang_2020_CVPR} version where half of the classes are biased toward color and the other half is biased toward gray where the dominant sensitive group in each target represents 95\% of the samples. 

\noindent\textbf{Metrics} Using Average accuracy on each sample is a misleading metric. This is because the test set could be biased toward some subgroups more than others. Therefore, the metric does not reflect how the model performs on \textit{all} subgroups. Methods, therefore, are evaluated in terms of Unbiased Accuracy \cite{hong2021unbiased}, which computes the accuracy of each subgroup separately and then returns the mean of accuracies, and Bias-Conflict \cite{hong2021unbiased}, which measures the accuracy on the minority subgroups. 

\noindent\textbf{Baselines} We report several baselines from prior work. First, we report "Bias-Contrastive and Bias-Balanced Learning" (BC + BB) \cite{hong2021unbiased}, which uses a contrastive learning framework to mitigate bias. Second, domain-independent (DI) \cite{Wang_2020_CVPR} uses an additional prediction head for each bias group. GroupDRO \cite{Sagawa2020DistributionallyRN} which optimizes a worst-group training loss combined with group-balanced resampling. Adversarial Learning (Adv) w/ uniform confusion \cite{Hoffman2015SimultaneousDT} introduces an adversarial loss that seeks to randomize the model's feature representation of $Y$ and $B$. Furthermore, we expand the benchmark by reporting the performance of sampling methods outlined in Section \ref{sec:methods}. Note that since Undersampling and Oversampling result in smaller/larger distributions per epoch than other methods, we adjust their number of training epochs such that the model sees the same number of data batches throughout the training process as other methods in our experiment. Refer to the Supplementary for further details. Finally, all reported methods are compared to a "vanilla" model trained on the original dataset distribution with a Cross-Entropy loss. All baselines are trained using the same encoder. Refer to the Supplementary for further details. 

\noindent\textbf{Results} in Table \ref{tb:main-results} demonstrates how BM is the most robust sampling method. It achieves strong performance on every dataset, unlike other sampling methods. For example, Undersampling while performing well on CelebA (likely because the task of predicting hair color is easy), performs considerably worse on every other dataset. Moreover, Oversampling, performs consistently worse than our method. This aligns with \cite{Wang_2020_CVPR} observation that neural nets tend to overfit when trained on oversampled distributions. Finally, while Upweighting maintains strong performance on CelebA and UTK-Face, it falls behind on CIFAR-S. This is because the method struggled to optimize, a property of Upweighting that has been noted and discussed in \cite{DBLP:conf/iclr/AnYZ21}. BM, consequently, improves over Upweighting on CIFAR-S by over $\%10$ on the Bias Conflict metric.

While our method is the only sampling method that shows consistently strong performance, vanilla sampling methods are surprisingly effective on some datasets. For example, Upweighting is effective on UTK-Face and CelebA, especially when compared to non-sampling methods. Moreover, Undersampling similarly is more effective than non-sampling methods on CelebA. This is evidence that when enough data is available and the task is "easy" enough, then simple Undersampling is effective. These results are important because they show how simply fixing the dataset distribution through sampling methods could be a strong baseline. We encourage future work, therefore, to add these sampling methods as part of their baselines. 

With respect to non-sampling methods, our method (BM) maintains strong comparable performance. This strong performance comes with \textbf{\textit{no}}  additional loss functions or any model modification which limits the hyper-parameter space and implementation complexity of our method compared to non-sampling methods. For example, BC+BB \cite{hong2021unbiased} requires a scalar parameter to optimize an additional loss. It also requires choosing a set of augmentation functions for its contrastive learning paradigm, which must be tuned according to the dataset. GroupDRO \cite{sagawa2019distributionally}, on the other hand, requires careful L2 regularization as well as tuning their "group adjustment"  hyper parameter. As a result, BM is faster and cheaper to train.

\subsection{The Effect of Each \texorpdfstring{$d_c$}{Lg} on Model Performance}
\label{sec:analysis_dy}
For each class, $c\in C$, Bias Mimicking returns a subsampled version of the dataset  $d_c \subset D$ where class $c$ samples are preserved, and the bias of $c$ is mimicked in the other classes. In this section, we investigate the effect of each $d_c$  on performance. We use the binary classification tasks in Section \ref{sec:main_results}. For each task, thus, we have two versions of the dataset $d_{c_1}, d_{c_2}$ where $c_1$ is the first class and $c_2$ is the second. We compare three different models performances: model (1) trained on only $d_{c_1}$, model (2) trained on only $d_{c_2}$, and finally, model (3) trained on both $(d_{c_1}, d_{c_2})$. The last version is the one used by our method. We use the Unbiased Accuracy Metric (UA). We also break down the metric into two versions: $\text{UA}_1$ where accuracy is averaged over class 1 subgroups, and $\text{UA}_2$ where accuracy is averaged over class 2 subgroups. Observe the results in Table \ref{tab:dyeffect}. Overall, the model trained on $(d_{c_1})$ performs better at predicting $c_1$ but worse at predicting $c_2$. We note the same trend with the model trained on $(d_{c_2})$ but in reverse. This disparity in performance harms the average Unbiased Accuracy (UA). However, a model trained on $(d_{c_1}, d_{c_2})$ balances both accuracies and achieves the best total Unbiased Accuracy (UA). These results emphasize the importance of each $d_c$ for good performance.

\begin{figure*}[t!]
    \centering
    \includegraphics[width=\linewidth]{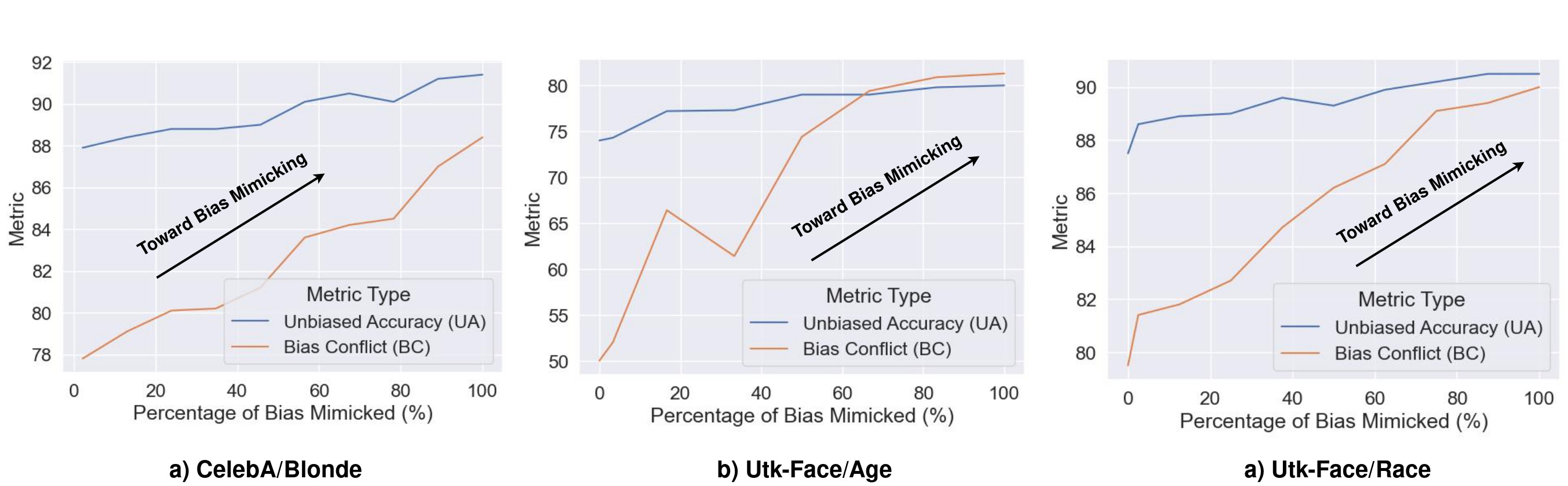}
    \caption{\textbf{Model Sensitivity to Bias Mimicking} We test our method's sensitivity to the Bias Mimicking condition in Equation \ref{eq:biasmimick}. To that end, we simulate multiple scenarios where we mimic the bias by $x\% \in \{0, 100\}$ (x-axis) such that $0\%$ represents no modification to the distribution and $100\%$ represents complete Bias Mimicking. We report results on the three binary classification benchmarks defined in Section \ref{sec:main_results}. Refer to Section \ref{sec:exp_analysis} for further details and discussion.} 
    \label{fig:main_figure}
    \vspace{-4mm}
\end{figure*}

\subsection{Model Sensitivity to the Mimicking Condition}
\label{sec:exp_analysis}

In this section, we test the model sensitivity to the mimicking condition outlined in Eq. \ref{eq:biasmimick}. To that end, we consider the binary classification benchmarks in Section \ref{sec:main_results}. These benchmarks have two classes, $c_1$ and $c_2$, and two bias groups, $b_1$ and $b_2$. Each class is biased toward one of the bias groups. We vary the mimicking process according to a percentage value where for value  $0\%$, the resulting $d_c$(s)  are the same as the original training distribution, and for value $x > 0\%$,   $d_c(s)$ are subsampled from the main training distribution such that the bias is mimicked by $x\%$. Note how, therefore, $100\%$ represents the full mimicking process as outlined in Eq. \ref{eq:biasmimick} whereas $x=50\%$ mimicks the bias by half. Formally, if $c_2$ needs to mimick $c_1$ and $c_1$ is biased toward $b_1$ then 

\begin{equation*}
    P_{d_{c_1}}(B = b_1 | Y=c_2 ) = \frac{1}{2} P_{d_{c_1}}(B = b_1 | Y=c_1 )
\end{equation*}

Observe the result in Figure \ref{fig:main_figure}. Note that as the percentage of bias Mimicked decreases, the BC and UA decrease as expected. This is because $P_D(Y | B) \neq P_D(Y)$ following proposition \ref{prop:1}.  More interestingly, note how the Bias Conflict (the accuracy over the minority subgroups) decreases much faster. The best performance is achieved when $x\%$ is indeed $100\%$ and thus $P_D(Y | B) = P_D(Y)$. From this analysis, we conclude that the Bias Mimicking condition is critical for good performance.

\section{Conclusion}
In this paper, we observed that hyper-parameter-free sampling methods for bias mitigation like Undersampling and Upweighting were missing from recent benchmarks. Therefore, we benchmarked these methods and concluded that some are surprisingly effective on many datasets. However, on some others, their performance significantly lagged behind non sampling methods. Motivated by this observation, we introduced a novel sampling method: Bias Mimicking. The method retained the simplicity of sampling methods while bridging the performance gap between sampling and non sampling methods. Furthermore, we extensively analyzed the behavior of Bias Mimicking, which emphasized the importance of our design. We hope our new method can reestablish sampling methods as a promising direction for bias mitigation. 

\smallskip\noindent\textbf{Limitations And Future Work} We demonstrated that dataset re-sampling methods are simple and effective tools in mitigating bias. However, we recognize that the explored bias scenarios in prior and our work are still limited. For example, current studies only consider mutually exclusive sensitive groups. Thus, a sample can belong to only one sensitive group at a time. How does relaxing this assumption, \ie intersectionality, impact bias? Finally, the dataset re-sampling methods presented in this work are effective at mitigating bias when enough samples from all dataset subgroups are available. However, it is unclear how these methods could handle bias scenarios when one class is completely biased by one sensitive group. Moreover, sampling methods require full knowledge of the bias groups labels at training time. Collecting annotations for these groups could be expensive. Future work could benefit from building more robust models independently of bias groups labels.
\smallskip

\noindent\textbf{Acknowledgements} This material is based upon work supported, in part, by DARPA under agreement number HR00112020054 and the National Science Foundation, including under Grant No.\ DBI-2134696. Any opinions, findings, and conclusions or recommendations expressed in this material are those of the author(s) and do not necessarily reflect the views of the supporting agencies.

\clearpage

{\small
\bibliographystyle{ieee_fullname}
\bibliography{egbib}
}
\appendix

\begin{table}[t!]
\sisetup{table-number-alignment=center}
\sisetup{
  table-align-uncertainty=true,
  separate-uncertainty=true,
  detect-all,
  detect-weight=true,
  detect-shape=true, 
  detect-mode=true,
}
\setlength{\tabcolsep}{0.3em}
\renewcommand{\arraystretch}{1.2}
\centering
\begin{tabular}{llS[table-format=2.1(2),detect-all,mode=text] S[table-format=2.1(2),detect-weight,mode=text]S[table-format=2.1(2),detect-all,mode=text]S[table-format=2.1(2),detect-all,mode=text]}
\toprule
 &  & {BM} & {BM+US} & {BM+UW} & {BM+OS} \\ \midrule
UTK-Face & UA & 79.7\pm0.4 & 79.2\pm1.0 & 79.9\pm0.2 & 79.3\pm0.2 \\
Age & BC & 79.1\pm2.3 & 77.6\pm0.9 & 77.5\pm1.7 & 78.7\pm1.6 \\ \midrule
UTK-Face & UA & 90.8\pm0.2 & 90.9\pm0.5 & 91.1\pm0.2 & 90.7\pm0.4 \\
Race & BC & 90.7\pm0.5 & 91.1\pm0.3 & 91.6\pm0.1 & 90.9\pm0.5 \\ \midrule
CelebA & UA & 90.8\pm0.4 & 91.1\pm0.2 & 91.1\pm0.4 & 91.1\pm0.1 \\
Blonde & BC & 87.1\pm0.6 & \underline{\tablenum{87.9\pm0.3}} & \underline{\tablenum{87.9\pm0.7}} & \underline{\tablenum{87.7\pm0.4}} \\ \midrule
CIFAR-S & UA & 91.6\pm0.1 & 91.7\pm0.1 & 91.8\pm0.0 & 91.6\pm0.2 \\
 & BC & 91.1\pm0.1 & 91.2\pm0.2 & 91.4\pm0.2 & 91.2\pm0.2\\
 \bottomrule
\end{tabular}
\caption{\textbf{Sampling for multi-class prediction head} compare the effects of using different sampling methods to train the multi-class prediction in our proposed method: Bias Mimicking. We underline results where sampling methods make significant improvements. Refer to Section \ref{sec:supp_bm_sampling} for discussion. }
\label{tb:supp_bm_with_sampling}
\vspace{-4mm}
\end{table}

\section{Sampling methods Impact on Multi-Class Classification Head}
\label{sec:supp_bm_sampling}
Bias Mimicking produces a binary version $d_c$ of the dataset $D$ for each class $c$. Each $d_c$ preserves class $c$ samples while undersampling each $c^{\prime}$ such that the bias within $c^{\prime}$ mimics that of $c$. A debiased feature representation is then learned by training a binary classifier for each $d_c$. When the training is done, using the scores from each binary predictor for inference is challenging. This is because each predictor is trained on a different distribution of the data, so the predictors are uncalibrated with respect to each other. Therefore, to perform inference, we train a multi-class prediction head using the learned feature representations and the original dataset distribution. Moreover, we prevent the gradients from flowing into the feature space since the original distribution is biased. Note that we rely on the assumption that the correlation between the target labels and bias labels are minimized in the feature space, and thus the linear layer is unlikely to relearn the bias. During our experiments outlined in Section 4, we note that this approach was sufficient to obtain competitive results. This section explores whether we can improve performance by using sampling methods to train the linear layer. To that end, observe results in Table \ref{tb:supp_bm_with_sampling}. We underline the rows where the sampling methods make improvements. We note that the sampling methods did not improve performance for three of the four benchmarks in our experiments. However, on CelebA, we note that the sampling methods marginally improved performance. We suspect this is because a small amount of the bias might be relearned when training the multi-class prediction head since the input distribution remains biased.

\begin{figure*}[ht!]
    \centering
    \includegraphics[width=\linewidth]{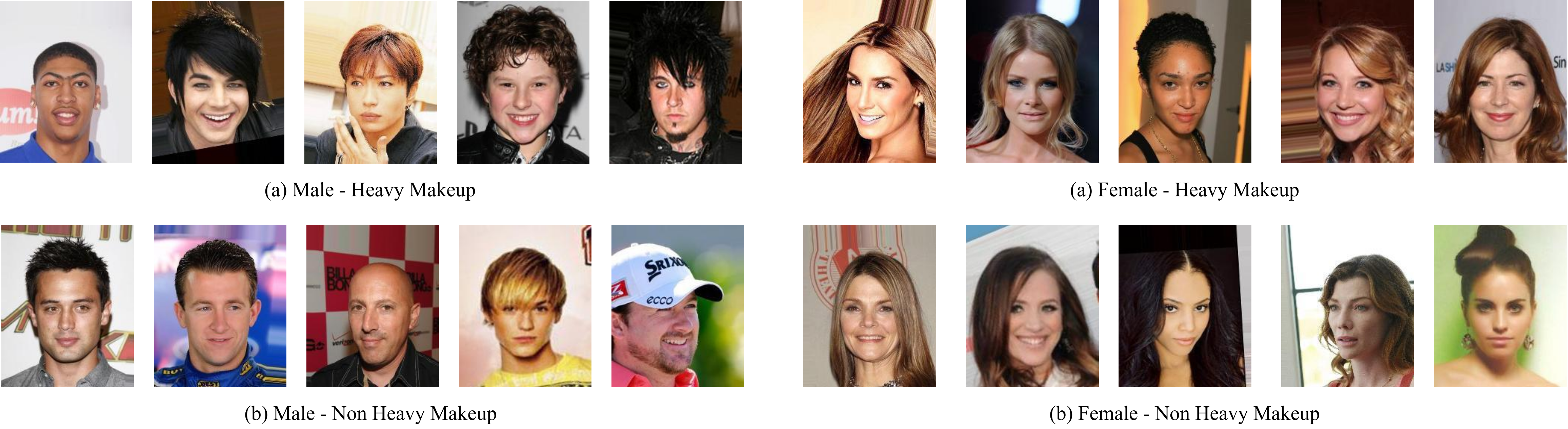}
    \caption{Randomly sampled images from the four subgroups: Female-Heavy Makeup, Female-Non-Heavy Makeup, Male-Heavy Makeup, and Male-Non-Heavy Makeup in CelebA dataset  \cite{liu2015faceattributes}. Note the that there is not a clearly differentiating signal for the attribute Heavy Makeup. Refer to Section \ref{sec:supp_heavy_makeup} for discussion.}
    \label{fig:supp_makeup}
\end{figure*}

\begin{table*}[t!]
\sisetup{table-number-alignment=center}
\label{tb:supp-binary-benchmark}
\sisetup{
  table-align-uncertainty=true,
  separate-uncertainty=true,
  detect-all,
  detect-weight=true,
  detect-shape=true, 
  detect-mode=true,
}
\setlength{\tabcolsep}{0.15em}
\renewcommand{\arraystretch}{1.2}
\centering
\begin{tabular}{llS[table-format=2.1(2),detect-all,mode=text] S[table-format=2.1(2),detect-weight,mode=text]S[table-format=2.1(2),detect-all,mode=text]S[table-format=2.1(2),detect-all,mode=text]S[table-format=2.1(2),detect-all,mode=text]|S[table-format=2.1(2),detect-all,mode=text]S[table-format=2.1(2),detect-all,mode=text]S[table-format=2.1(2),detect-all,mode=text]S[table-format=2.1(2),detect-all,mode=text]S[table-format=2.1(2),detect-all,mode=text]S[table-format=2.1(2),detect-all,mode=text]}
 \toprule
 &  & \multicolumn{5}{c|}{Nonsampling Methods} & 
 \multicolumn{5}{c}{Sampling Methods} \\ \toprule
 &  & {Vanilla} & {Adv \cite{Hoffman2015SimultaneousDT}} & {G-DRO \cite{Sagawa2020DistributionallyRN}} & {DI\cite{Wang_2020_CVPR}} & {BC+BB}\cite{hong2021unbiased} & {OS \cite{Wang_2020_CVPR}} & {UW \cite{pmlr-v97-byrd19a}} & {US\cite{Japkowicz02theclass}} & {BM} & {BM + OS} \\ \midrule
\multirow{2}{*}{\begin{tabular}[c]{@{}l@{}}CelebA\\ Smiling\end{tabular}} & UA & 90.9\pm 0.1 & 90.7\pm0.2 & 93.2\pm0.1 & 92.0\pm0.1 & 92.4\pm0.1 & 92.4\pm0.3 & 92.8\pm0.0 & 92.6\pm0.1 & 92.7\pm0.1 & 92.7\pm0.1 \\
 & BC & 84.3\pm 0.2 & 84.7\pm0.4 & 92.2\pm0.1 & 91.3\pm0.2 & 92.6\pm0.1 & 91.5\pm0.2 & 92.4\pm0.2 & 92.1\pm0.2 & 92.3\pm0.2 & 92.2\pm0.1 \\
\multirow{2}{*}{\begin{tabular}[c]{@{}l@{}}CelebA\\ Black Hair\end{tabular}} & UA & 86.3\pm0.7 & 87.1\pm0.3 & 88.5\pm0.2 & 86.7\pm0.7 & 87.7\pm0.1 & 87.6\pm0.3 & 88.5\pm0.1 & 88.4\pm0.2 & 87.6\pm0.7 & 88.5\pm0.1 \\
 & BC & 82.7\pm0.6 & 83.4\pm0.5 & 88.3\pm0.4 & 86.6\pm1.2 & 86.6\pm0.3 & 85.6\pm0.6 & 88.0\pm0.2 & 87.3\pm0.1 & 87.8\pm1.3 & 88.5\pm0.7 \\ \midrule
\multirow{2}{*}{Average} & UA & 88.6\pm0.4 & 88.9\pm0.2 & \bfseries 90.8\pm0.1 & 89.3\pm0.4 & 90.0\pm0.1 & 90.0\pm0.3 & \bfseries90.6\pm0.1 & \bfseries90.5\pm0.1 & \bfseries90.2\pm0.4 & \bfseries90.6\pm0.1 \\
 & BC & 83.5\pm0.4 & 84.0\pm0.4 & \bfseries 90.2\pm0.2 & 88.9\pm0.7 & 89.6\pm0.2 & 88.5\pm0.4 & \bfseries90.2\pm0.2 & 89.7\pm0.1 & \bfseries90.0\pm0.7 & \bfseries90.3\pm0.4\\
 \bottomrule
 \end{tabular}
\caption{Expanded benchmarks from the CelebA dataset \cite{liu2015faceattributes}. Refer to Section \ref{sec:supp_more_celeba} for discussion.}
\label{tb:supp_more_celeba}
\vspace{-0.1mm}
\end{table*}

\section{Heavy Makeup Benchmark}
\label{sec:supp_heavy_makeup}
Prior work \cite{hong2021unbiased} uses the Heavy Makeup binary attribute prediction task from CelebA  \cite{liu2015faceattributes} as a benchmark for bias mitigation, where Gender is the sensitive attribute. In this experiment, Heavy Makeup's attribute is biased toward the sensitive group: Female. We note that the notion of "Heavy Makeup" is quite subjective. The attribute labels may vary significantly according to cultural elements, lighting conditions, and camera pose considerations. Thus, we expect a fair amount of label noise, \ie, inconsistency with label assignment. We document this problem in a Quantitative and Qualitative analysis below. 

\smallskip\noindent\textbf{Quantitative Analysis:} We randomly select a total of 200 pairs of positive and negative images. We ensure the samples are balanced among the four possible pairings, \ie, (Heavy Makeup-Male, Non Heavy Makeup-Female), (Heavy Makeup-Female, Non Heavy Makeup-Male), (Heavy Makeup-Male, Non Heavy Makeup-Male), (Heavy Makeup-Female, Non Heavy Makeup-Female). We asked three independent annotators to label which image in the pair is wearing "Heavy Makeup". Then, we calculate the percentage of disagreement between the three annotators and the ground truth labels in the dataset. We note that $32.3\% \pm 0.02$ of the time, the annotators on average disagreed with the ground truth. The noise is further amplified when the test set used in \cite{hong2021unbiased} is examined. In particular, Male-Heavy Make up (an under-represented subgroup) only contains 9 testing samples. We could not visually determine whether 4 out of these 9 images fall under Heavy Makeup. Out of the 5 remaining images, 3 are of the same person from different angles. Thus, given the noise in the training set, the small size of the under-represented group in the test set, and its label noise, we conclude that results from this benchmark will not be reliable and exclude it from our experiments. 

\smallskip\noindent\textbf{Qualitative Analysis:} We sample random 5 images from the following subgroups: Female-Heavy Makeup, Female-Non-Heavy Makeup, Male-Heavy Makeup, and Male-Non-Heavy Makeup (Fig \ref{fig:supp_makeup}). It is clear from the Figure that there is no firm agreement about the definition of Heavy Makeup.

\section{Additional Benchmarks}
\label{sec:supp_more_celeba}
In Section \ref{sec:supp_heavy_makeup}, we note that the CelebA attribute HeavyMakeup usually used in assessing model bias in prior work \cite{hong2021unbiased,Tartaglione_2021_CVPR} is a noisy attribute, \ie, labels are inconsistent. Therefore, we choose not to use it in our experiments. Alternatively, we provide results on additional attributes where labels are more likely to be consistent. To that end, we choose to classify the attributes: Smiling and Black Hair, where Gender is the bias variable. The original distribution of each attribute is not sufficiently biased with respect to Gender to note any significant change in performance. Thus, we subsample each distribution to ensure that each attribute is biased toward Gender. We provide the splits for the resulting distributions in the attached code base. Refer for Table \ref{tb:supp_more_celeba} for results.

Note that our method Bias Mimicking performance marginally lags behind other methods when predicting "Black Hair" attribute. However, when the multi-class prediction layer is trained with an oversampled distribution (BM + OS), then the gap is bridged. This is consistent with the observation in Table \ref{tb:supp_bm_with_sampling} where oversampling marginally improves our method performance on CelebA. These observations indicate that on some benchmarks, a small amount of the bias might be relearned through the multi-class prediction head. To ensure that this bias is mitigated, it is sufficient to oversample the input distribution. Moreover, since oversampling the input distribution does not change performance on other datasets as indicated in Table \ref{tb:supp_bm_with_sampling}, we recommend that the input distribution for the multi-class prediction head is oversampled to ensure the best performance.

Overall, (BM + OS) performs comparably to sampling and nonsampling methods. This is consistent with our results on CelebA dataset in Section 4 of the main paper where we predict "Blonde Hair". More concretely, Undersampling performs comparably and sometimes better than nonsampling methods. This is reaffirming that predicting attributes on CelebA is relatively an easy task that dropping samples to balance subgroup distribution is sufficient to mitigate bias. However, as discussed in Section 4 of the main paper, vanilla sampling methods (Undersampling, Upweighting, Oversampling) perform poorly on some datasets. For example, as we note in Table 1 in the main paper, Undersampling performs considerably worse than nonsampling methods on the Utk-Face dataset as well the CIFAR-S dataset. Moreover, Upweighting performs substantially worse on CIFAR-S. Finally, Oversampling performs consistently worse on every benchmark. However, only our method, Bias Mimicking, manages to maintain competitive performance with respect to nonsampling methods on all datasets. 

\begin{table}[]
\renewcommand{\arraystretch}{1.0}
\begin{tabular}{l|lll}
\toprule
 & \begin{tabular}[c]{@{}l@{}}Learning \\ Rate\end{tabular} & \begin{tabular}[c]{@{}l@{}}Weight \\ Reg\end{tabular} & \begin{tabular}[c]{@{}l@{}}Group \\ Adjustment\end{tabular} \\ \midrule
UTK-Face Age & 0.001 & 0.01 & 4 \\
UTK-Face Race & 0.001 & 0.001 & 4 \\ \midrule
CelebA Blonde & 0.001 & 0.1 & 3 \\ 
CelebA Smiling & 0.0001 & 0.01 & 2 \\
CelebA Black Hair & 0.0001 & 0.01 & 3 \\\midrule
CIFAR-S & 0.01 & 0.01 & 5 \\ \bottomrule
\end{tabular}
\caption{Hyperparameters used for GroupDRO \cite{Sagawa2020DistributionallyRN}. Refer to Section \ref{sec:supp_hyper_param} for further discussion.}
\label{table:hyper_param_dro}
\end{table}

\begin{table}[t!]
\renewcommand{\arraystretch}{1.0}
\begin{tabular}{l|lll}
\toprule
 & US \cite{Japkowicz02theclass} & OS \cite{Wang_2020_CVPR}& Other Methods \\ \midrule
UTK-Face Age & 400 & 7 & 20 \\ 
UTK-Face Race & 120 & 10 & 20 \\ \midrule
CelebA Blonde & 170 & 4 & 10 \\ 
CelebA Black Hair & 40 & 5 & 10 \\
CelebA Smiling & 30 & 5 & 10 \\ \midrule
CIFAR-S & 2000 & 100 & 200 \\ \bottomrule
\end{tabular}
\caption{Number of Epochs used to train each method. Refer to Section \ref{sec:supp_hyper_param} for further discussion.}
\label{table:epochs_num}
\end{table}

\section{Model and Hyper-parameters Details}
\label{sec:supp_hyper_param}
We test bias Mimicking on six benchmarks. Three Binary Classification tasks on CelebA \cite{liu2015faceattributes}, namely, Blonde, Black Hair, and Smiling, Two Binary Classification tasks on UTK-Face \cite{zhifei2017cvpr}, namely Race and Age and one multi-class task CIFAR-S. We provide further info below.

\smallskip\noindent\textbf{Optimization} Following \cite{hong2021unbiased}, we use ADAM \cite{adam_optimizer} optimizer with learning rate $0.0001$ on CelebA and UTK-Face. For CIFAR-S, following \cite{Wang_2020_CVPR}, we use SGD with learning rate $0.1$. GroupDRO \cite{Sagawa2020DistributionallyRN}, however, has not been tuned before on the benchmarks in our study. Even for CelebA Blonde, the method was not tuned on the more challenging split in this study. Therefore, we grid search the learning rate/weight regularization/group adjustment and choose the best over the validation set. Refer to Table \ref{table:hyper_param_dro} for our final choices. With respect to BC+BB, the method was not benchmarked on CIFAR-S.  Therefore, we run a grid search over the method's hyperparameters and choose $\alpha = 1.0$, $\gamma = 10$. Finally, as discussed in Section 4, UW struggles to optimize over CIFAR-S with learning rate $0.1$. Therefore, we tune the learning rate and we find that $0.0001$ to work the best over the validation set. 

\smallskip\noindent\textbf{Total Number of Epochs} As noted in Section 4.1 in the paper, a model trained with Undersampling sees fewer iterations than baselines per epoch and a model trained with Oversampling sees more iterations per epoch. Therefore, we adjust the number of epochs for both methods such that the total number of iterations seen by the model is the same across all methods tested in our experiments. Refer to Table \ref{table:epochs_num} for a breakdown of the total number of epochs used to train each method.

\smallskip\noindent\textbf{Augmentations:} For all benchmarks, we augment the input images with a horizental flip. BC+BB \cite{hong2021unbiased} uses extra augmentation functions. Refer to \cite{hong2021unbiased} for further details. 

\smallskip\noindent\textbf{Splits:} Note on CelebA, unlike \cite{hong2021unbiased}, we use CelebA validation set for validation and test set for testing rather than using the validation set for testing and a split of the training set for validation.

\begin{table}[t!]
\sisetup{table-number-alignment=center}
\sisetup{
  table-align-uncertainty=true,
  separate-uncertainty=true,
  detect-all,
  detect-weight=true,
  detect-shape=true, 
  detect-mode=true,
}
\setlength{\tabcolsep}{1.0em}
\renewcommand{\arraystretch}{0.8}
\centering
\begin{tabular}{l|S[table-format=2.1,detect-all]S[table-format=2.1,detect-all]S[table-format=2.1,detect-all]S[table-format=2.1,detect-all]}
\toprule
 & {100\%} & {75\%} & {50\%} & {25\%} \\ \midrule
UTK-Face Age & \bfseries 79.7 & 78.9 & 78.0 & 76.7 \\
UTK-Face Race & \bfseries 90.8 & 90.6 & 89.9 & 88.3 \\
CelebA Blonde & \bfseries 90.8 & 90.3 & 90.1 & 89.9 \\
\bottomrule
\end{tabular}
\vspace{-3.0mm}
\caption{Comparing the performance of our method's Unbiased Accuracy (UA) where we use $x\%$ of the linear program solution. Refer to \ref{supp:linear_program_solution} for discussion.}
\label{table:linear_program}
\vspace{-2.0mm}
\end{table}

\section{Effect of the Linear Program Constraint}
\label{supp:linear_program_solution}
As discussed in Section 3 in the paper, We use a linear program to determine how the training distribution is subsampled to mimick the bias. The program is constrained such that the resulting distributions preserve the most number of samples because fewer retained samples may compromise the model performance. To verify this the importance of this step, we train our model using distributions that maintain  x\% of the Linear Program solution where $x < 100$. Note the results in Table \ref{table:linear_program}. Note how the performance drops emphasizing the importance of this constraint.

\end{document}